\documentclass[%
 reprint,
 amsmath,amssymb,
 aps,
]{revtex4-2}
\usepackage{enumitem}
\usepackage{hyperref}% add hypertext capabilities
\usepackage[capitalise]{cleveref}
\usepackage{graphicx}% Include figure files
\usepackage{dcolumn}% Align table columns on decimal point
\usepackage{bm}% bold math
\usepackage{tabularx}
\usepackage{booktabs}
\usepackage{array}

\usepackage{placeins}

\newcolumntype{C}[1]{>{\centering\arraybackslash}p{#1}}

\begin{document}

\preprint{APS/123-QED}
\title{Moderate Adaptive Linear Units (MoLU)}
% \thanks{A footnote to the article title}%

\author{\textbf{Hankyul Koh}}
\altaffiliation[]{Physics, College of Natural Sciences Dept. of Physics \& Astronomy, Seoul National University, South Korea\\}
\email{physics113@snu.ac.kr}

\author{\textbf{Joon-Hyuk Ko}}
\altaffiliation[]{Physics, College of Natural Sciences Dept. of Physics \& Astronomy, Seoul National University, South Korea\\}
\email{jhko725@snu.ac.kr}
\affiliation{Seoul National University   \space Seoul National University}

\author{\textbf{Wonho Jhe}}
\altaffiliation[]{Physics, College of Natural Sciences Dept. of Physics \& Astronomy, Seoul National University, South Korea\\}
\email{whjhe@snu.ac.kr}
\affiliation{Seoul National University}
% \collaboration{MUSO Collaboration}%\noaffiliation

% \author{Charlie Author}
%  \homepage{http://www.Second.institution.edu/~Charlie.Author}
% \affiliation{
%  Second institution and/or address\\
%  This line break forced% with \\
% }%
% \affiliation{
%  Third institution, the second for Charlie Author
% }%

\date{\today}% It is always \today, today,
             %  but any date may be explicitly specified

\begin{abstract}
We propose the \textit{Moderate Adaptive Linear Unit (MoLU)}, a novel activation function for deep neural networks, defined analytically as: \(f(x)=x \cdot \frac{1}{2} (1+tanh(x))\). MoLU combines mathematical elegance with empirical effectiveness, exhibiting superior performance in terms of prediction accuracy, convergence speed, and computational efficiency. Due to its \(C^\infty\)-smoothness, i.e. infinite differentiability and analyticity, MoLU is expected to mitigate issues such as vanishing or exploding gradients, making it suitable for a broad range of architectures and applications, including large language models (LLMs), Neural Ordinary Differential Equations (Neural ODEs) \cite{neuralode}, Physics-Informed Neural Networks (PINNs) \cite{pinn}, and Convolutional Neural Networks (CNNs). Empirical evaluations show that MoLU consistently achieves faster convergence and improved final accuracy relative to widely used activation functions such as GeLU \cite{gelu}, SiLU\cite{silu}, and Mish\cite{mish}. These properties position MoLU as a promising and robust candidate for general-purpose activation across diverse deep learning paradigms. 
% \begin{description}
% \item[Usage]
% Secondary publications and information retrieval purposes.
% \item[Structure]
% You may use the \texttt{description} environment to structure your abstract;
% use the optional argument of the \verb+\item+ command to give the category of each item. 
% \end{description}
\end{abstract}

%\keywords{Suggested keywords}%Use showkeys class option if keyword
                              %display desired
\maketitle

%\tableofcontents

\section{\label{sec:level1}Introduction}

In the context of Neural Ordinary Differential Equations (Neural ODEs), early studies typically employed activation functions such as the Exponential Linear Unit (ELU) \cite{elu} or the hyperbolic tangent (Tanh) function, primarily due to their differentiability across the entire real domain. Despite this, Neural ODEs often exhibited limited predictive performance—not only on long time-series data but even on relatively short sequences. This underperformance was commonly attributed to the inherent limitations of the Neural ODE framework itself. Moreover, a critical drawback was the prolonged training time required to reach convergence. Our previously proposed homotopy-based training method \cite{homotopy} significantly improved both the convergence speed and predictive accuracy of Neural ODEs. Building upon this success, we sought further enhancement by focusing on a more fundamental component of the model—the activation function itself.\\
 In recent work, the Gaussian Error Linear Unit (GeLU) has been introduced in the context of Neural ODEs \cite{stiff} and was also employed in our prior work, as it outperformed Tanh in various tasks. However, through extensive evaluation, we observed that our newly proposed activation function outperforms GeLU in both accuracy and training efficiency. Furthermore, we demonstrate that our activation function achieves consistently superior performance in image classification tasks, including benchmark datasets such as MNIST \cite{mnist} and CIFAR-10 \cite{cifar}, significantly surpassing existing activation functions in both accuracy and training efficiency.
 
 % Gradient descent, which is mainly used to update parameters in the neural network, is associated with activation function, as the layer gets deeper, the problem of gradient exploding or gradient vanishing frequently occurs. The most commonly used Sigmoid or Tanh functions in the neural network have slope values of 0 to 0.3 for Sigmoid and 0 to 1 for Tanh, so the deeper the layer, the smaller the gradient becomes. ReLU came out to solve this problem. Since the slope of the ReLU is 1 when \(x>0\), it solved the gradient vanishing problem. \\

\section{\label{sec:level2}MoLU Formulation}
\subsection{MoLU}
\vskip -0.1in
The MoLU is defined as below in Eq.~(\ref{eq:one}).

\begin{equation}
\label{eq:one}
MoLU = x \times \frac{1}{2}(1+\tanh(x))
\end{equation}

\begin{figure}[!ht]
\includegraphics[width=0.45\textwidth]{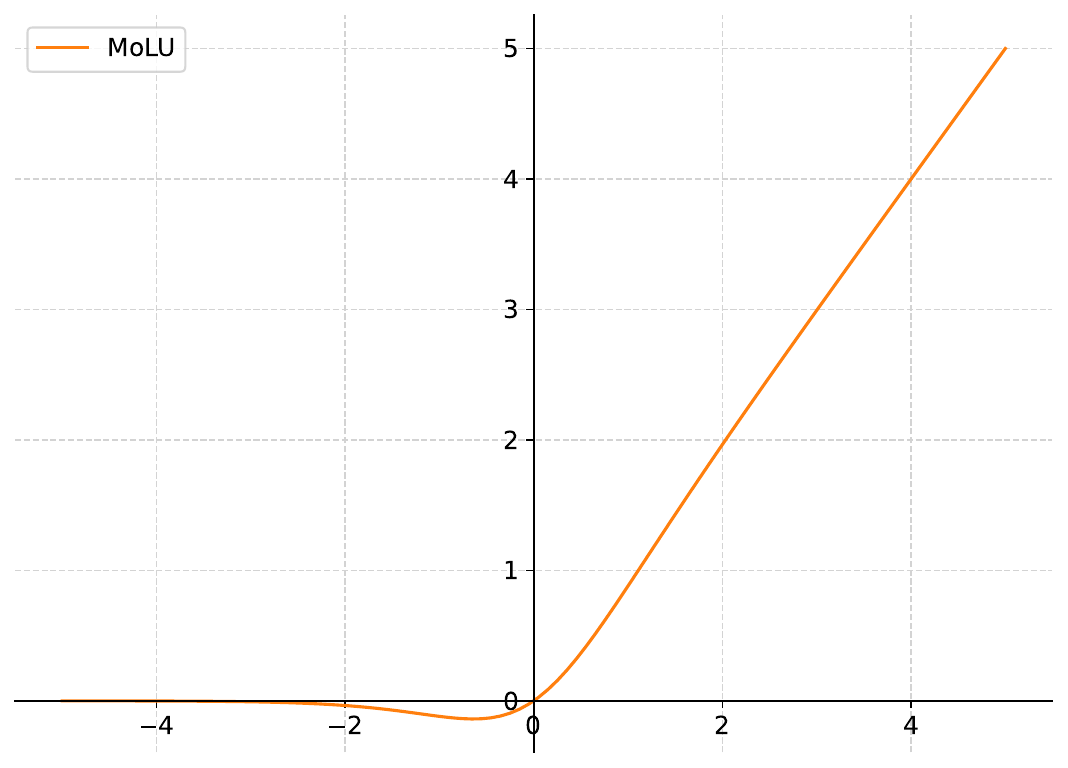}
\caption{\label{fig:MoLU} MoLU (Moderate Adaptive Linear Units).}
\end{figure}

\FloatBarrier

% \vskip -0.1in
\begin{figure*}[ht]
    \centering
    \includegraphics[width = 1.0\textwidth]{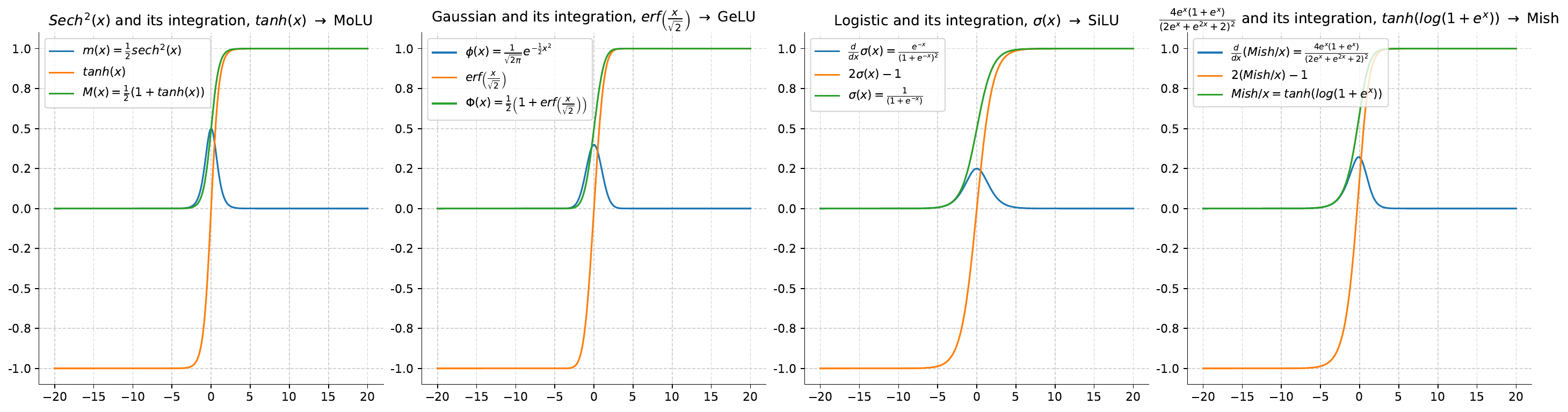}
    % \vskip -0.1in
    \caption{Tendencies. Three-step decomposition of activation functions: each panel illustrates the distribution \textbf{(blue)}, its cumulative distribution function form \textbf{(orange)}, and the affine-transformed cumulative distribution function mapping $(-1, 1)$ to $(0, 1)$ \textbf{(green)} for four representative cases MoLU, GeLU, SiLU, and Mish.}
\label{fig: tendencies}
\end{figure*}

\FloatBarrier

\subsection{Motivation}
We observed that activation functions of the form where an input is multiplied by a scaled and translated integral of a bell-shaped distribution, resulting in a logistic-like function with range confined to \((0, 1)\), tend to exhibit strong learning performance in practice. 

Based on our findings, activation functions that yield strong learning performance can be systematically constructed through the following steps:

\vskip 0.1in
\begin{enumerate}
    \item \textbf{Identify a suitable distribution}, denoted \(\phi(x)\) that satisfies the following conditions:
    \begin{enumerate}[label=(\roman*)]
        \item The domain is real-valued and range is positive real-valued, that is,
        \[
        x \in \mathbb{R}, \quad \phi(x) \in \mathbb{R}^+, \quad \text{with} \quad \lim_{x \to \pm \infty} \phi(x) = 0.
        \]
        \item The total area under the distribution is unity:
        \[
        \int_{-\infty}^{\infty} \phi(x)\, dx = 1.
        \]
        The center of the domain does not necessarily have to be located at the origin.
    \end{enumerate}

    \item \textbf{Integrate the distribution} to obtain its cumulative distribution function \( \Phi(x) \). If the range of \( \Phi(x) \) lies in \((-1, 1)\), apply an appropriate affine transformation to map it to the interval \((0, 1)\). This transformation is essential for proper activation behavior.

    \item \textbf{Construct the activation function} by multiplying the cumulative distribition function with the input \( x \):
    \[
    A(x) = x \cdot \Phi(x),
    \]
    where \( A(x) \) denotes the final activation function. This formulation combines the adaptive shaping of the probability distribution with the linear behavior near the origin, yielding both expressiveness and training stability.
\end{enumerate}

\subsection{Comparison with other activation functions}

Widely used activation functions such as GeLU and SiLU, which are prevalent in general deep learning frameworks, along with the MoLU introduced in this work, can be systematically formulated following the aforementioned three-step framework: (1) a bell-shaped distribution, (2) a logistic-like cumulative distribution function, and (3) the final activation function.
\vskip 0.1in

\textbf{Bell-shaped Distribution}
\begin{equation}
\begin{aligned}
\phi(x) &= \frac{1}{\sqrt{2\pi}} e^{-\frac{1}{2} x^2}, \quad \quad x \in (-\infty, \infty) \\
\sigma(x) &= \frac{e^x}{1 + e^x},  \ \ \ \quad  \quad \quad x \in (-\infty, \infty) \\
m(x) &= \frac{1}{2} \mathrm{sech}^2 x, \  \quad  \quad \quad x \in (-\infty, \infty)
\end{aligned}
\label{eqn: bell-shaped}
\end{equation}

\textbf{Logistic-like Functions}
\begin{equation}
\begin{aligned}
\Phi(x) &= \frac{1}{2} \left( 1 + \mathrm{erf}\left(\frac{x}{2}\right) \right), \quad \Phi(x) \in (0,1) \\
\sigma(x) &= \frac{1}{1 + e^{-x}}, \quad \quad \quad \quad \quad \sigma(x) \in (0,1) \\
M(x) &= \frac{1}{2} \left( 1 + \tanh(x) \right), \ \quad M(x) \in (0,1)
\end{aligned}
\label{eqn: logistic-like}
\end{equation}

\textbf{Activation functions}
\begin{equation}
\begin{aligned}
\mathrm{GeLU} &:= x \cdot \Phi(x) = x \cdot \frac{1}{2} \left( 1 + \mathrm{erf}\left(\frac{x}{\sqrt{2}}\right) \right), \\
\mathrm{SiLU} &:= x \cdot \sigma(x) = x \cdot \frac{1}{1 + e^{-x}}, \\
\mathrm{MoLU} &:= x \cdot M(x) = x \cdot \frac{1}{2} (1 + \tanh(x)).
\end{aligned}
\label{eqn: activations}
\end{equation}

% \begin{figure}[!ht]
% \includegraphics[width=0.45\textwidth]{figures/act_fcs.pdf}

% \caption{\label{fig:act funcs}Activation functions.}
% \end{figure}

% \FloatBarrier

\begin{figure*}[ht]
    \centering
    \includegraphics[width = 1.0\textwidth]{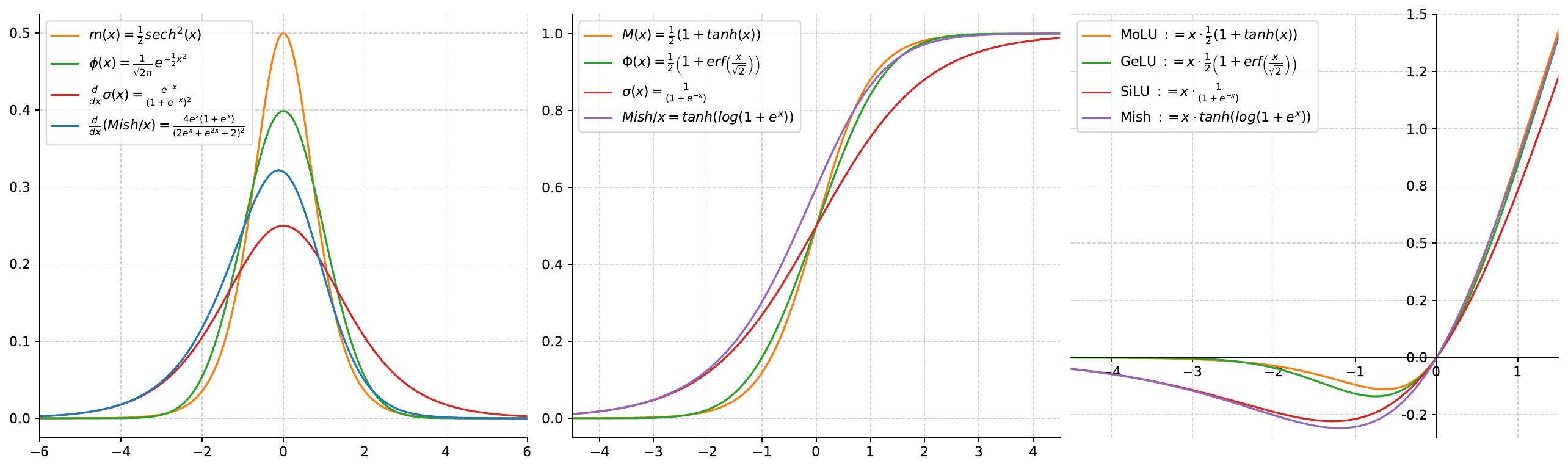}
    % \vskip -0.1in
    \caption{Activation functions. Comparison of the distribution functions \textbf{(left)}, their logistic-like functions \textbf{(middle)}, and the resulting activation functions \textbf{(right)} for MoLU, GeLU, SiLU, and Mish. The left panel shows bell-shaped probability density functions used as base functions, the middle shows their corresponding cumulative distribution functions scaled to $(0,1)$, and the right shows the final activation functions, which determine the nonlinearity applied in neural networks.}
\label{fig: activations}
\end{figure*}

To investigate this hypothesis more systematically, we conducted experiments using a range of activation functions, including GeLU, SiLU, Mish, ELU, Tanh, MoLU, as well as those derived from the student’s t-distribution with various degrees of freedom \(\nu\). Interestingly, the definition of MoLU, given by \(x \cdot \frac{1}{2} (1 + \tanh(x))\), is analytically equivalent to Swish \cite{swish}, defined as \(x \cdot \frac{1}{1 + e^{- \beta x}}\), when the parameter \(\beta=2\). However, despite this equivalence in functional form, MoLU demonstrated superior performance not only in terms of training loss, but also in training efficiency, significantly outperforming Swish with \(\beta=2\) and other activation functions in empirical evaluations. As is well known, when \(\nu \rightarrow \infty\), the student's t-distribution converges to the normal distribution, which underlies the GeLU. In the case of \(\nu=1\), the student’s t-distribution reduces to the Cauchy distribution. When this distribution is integrated and appropriately translated, it can be transformed into a logistic-like function whose values lie within the interval \((0, 1)\) over the entire real domain. However, when this cumulative function is multiplied by the input \(x\), the resulting activation function asymptotically approaches a nonzero constant, specifically \(-1/\pi\), as \(x \rightarrow - \infty\). This non-vanishing behavior prevents the loss function from effectively converging to a correct local minimum and, in practice, leads to poor training performance. For \(\nu=2\), activation functions constructed via the above method yield successful learning behavior. However, from \(\nu \geq 3\), although the distribution remains well-behaved, the resulting expressions become algebraically more complex, increasing computational overhead. This complexity negatively impacts both training speed and overall learning performance. The probability density function and cumulative distribution functions of the student’s t-distribution with \(\nu=2\) and \(\nu=3\) are given as follows:
\vskip -0.2in
\begin{equation}
    \begin{split}
        & \text{PDF}: \ \frac{1}{2\sqrt{2} \left( 1+ \frac{t^{2}}{2} \right)^{3/2}}, \quad \frac{2}{\pi \sqrt{3} \left( 1+ \frac{t^{2}}{3} \right)^{2}} \\
        & \text{CDF}: \  \frac{1}{2} + \frac{t/2\sqrt{2}}{ \sqrt{1+\frac{t^{2}}{2}}}, \quad \frac{1}{2} + \frac{1}{\pi} \left[ \frac{\left(\frac{t}{\sqrt{3}} \right)}{\left(1+\frac{t^{2}}{3} \right)} + tan^{-1}\left( \frac{t}{\sqrt{3}} \right) \right].
    \end{split}
\label{eqn: nu2}
\end{equation}

\subsection{Why do such forms of activation functions enable effective data prediction?}
To gain intuition, consider first the classical least-squares method, a simple yet powerful mathematical tool. This foundational concept offers insight into why more complex models like neural networks can achieve remarkably effective data fitting. 
Let us begin with the most elementary case: fitting a set of data points using a straight line via the least-squares method. Mathemetically, the best-fit line can be expressed in the parametric form

\vskip -0.2in
\begin{equation}
\mathbf{x} = \mathbf{A} + \lambda \mathbf{\hat{u}}.
\label{eqn: line fitting}
\end{equation}

where \(A\) is a point lying on the line, \(\mathbf{\hat{u}}\) is a directional vector, and \(\lambda \in \mathbb{R}\) is a scalar parameter, analogous to a weight in a neural network. In this formulation, the optimal parameters --- such as the slope and intercept --- are obtained by minimizing the sum of squared residuals between the predicted and observed values.
When an additional input variable in introduced, the fitting problem generalizes from a line to a plane. The parametric equation becomes

\vskip -0.1in
\begin{equation}
\mathbf{x} = \mathbf{A} + \lambda \mathbf{\hat{u}} + \mu \mathbf{\hat{v}}.
\label{eqn: plane fitting}
\end{equation}

where \(\mathbf{\hat{v}}\) is a directional vector, which is not parallel to \(\hat{u}\), and \(\mu \in \mathbb{R}\) is its associated parameter. As the dimensionality of the input increases, the number of fitting parameters grows accordingly, thus complicating the optimization process. In this context, the term dimensionality refers to the number of nodes within the layer. An increase in dimensionality corresponds to an increase in the number of nodes, which in turn signifies the addition of new directional vectors in the parametric space.

This observation becomes particularly insightful when applied to neural networks. In a typical feedforward layer, input data is linearly transformed through a combination of weights and biases, producing a set of affine expressions that serve as the domain for the activation functions. That is, for each node (or neuron), we compute

\vskip -0.2in
\begin{equation}
\begin{split}
   \text{Input:} &\quad [\mathrm{input}_i], \quad i=1, 2, 3, ..., n \\
    \rightarrow &\quad \omega_{\alpha} \times [\mathrm{input}_i] + b_{\alpha} = [x_{\alpha i}] \\
    \rightarrow &\quad \sigma(x_{\alpha i}) = [y_{\alpha i}] \\
    \rightarrow &\quad [y_{\alpha i}] \times w_{\alpha + n_{\mathrm{nodes}}} \\
    \rightarrow &\quad \sum_{\alpha} \left( [y_{\alpha i}] \times w_{\alpha + n_{\mathrm{nodes}}} \right) + b_{\alpha + n_{\mathrm{nodes}}} \\
    &\quad = [\mathrm{predicted}_i].
\end{split}
\label{eqn:neuralnetflow}
\end{equation}

and then evaluate \(\sigma([x_i])\), where n is the number of data, \(\alpha\) is the number of nodes in hidden layer, \(\sigma(x)\) is the activation function. 
Importantly, many activation functions exhibit distinct behavior across different regions of their domain. For instance, in the case of the ReLU function, values below zero are mapped to zero. This has a striking implication: the corresponding weight vector (or directional component) becomes effectively inactive. In geometric terms, rather than fitting a plane (or a higher-dimensional manifold), the model locally behaves as if fitting a line or a simpler surface.
Thus, the activation function serves as a nonlinear gate that sparsifies the parameter space. It effectively reduces the dimensionality of the local fitting problem by suppressing certain directions --- those associated with negative domain values. This implicit simplification turns an otherwise difficult high-dimensional fitting task into a collection of simpler subproblems, often resulting in better generalization and computational tractability.
Alternatively, this phenomenon can be interpreted from a statistical perspective. The nodes in a hidden layer may be regarded as an ensemble of randomly sampled experiments or measurements, each contributing to the approximation of the observed data during forward propagation. In this view, the process of predicting the output can be understood as fitting the observed data by randomly sampling across the various domain intervals of the activation function.
In activation functions of the form considered here, the positive domain behaves nearly as an identity transformation. This facilitates the propagation of values that closely resemble the true measurements, thereby enhancing the fidelity of the predicted outputs. Conversely, the negative domain, which maps inputs to values near zero, may be interpreted as introducing a small statistical perturbation. 
In summary, activation functions do more that introduce nonlinearity --- they act as regulators that dynamically modulate the model's degrees of freedom. This allows neural networks to approximate complex functions while avoiding overfitting and maintaining efficient representation, analogous to selectively fitting lower-dimensional affine subspaces depending on the data structure.

Let us now examine the backpropagation process. Consider estimating \(\omega_{\alpha=1}\), the scaling factor for the first node, which also serves as the slope in a local linear fitting. To perform gradient descent, we differentiate the sum of squared residuals with respect to this parameter using the chain rule, as shown below.

\begin{equation}
\begin{split}
&\mathcal{L} =  \Sigma_i(\mathrm{observed_i}-\mathrm{predicted}_i)^{2}, \quad i=1, 2, 3, ..., n \\
&\frac{\partial \mathcal{L}}{\partial \omega_{\alpha}} = \frac{\partial \mathcal{L}}{\partial \mathrm{predicted_i}} \times \frac{\partial \mathrm{predicted_i}}{\partial y_{\alpha i}} \times \frac{dy_{\alpha i}}{dx_{\alpha i}} \times \frac{dx_{\alpha i}}{d \omega_{\alpha}}.
\end{split}
\label{eqn:backpropagation}
\end{equation}

In the gradient descent formulation in Eq.~(\ref{eqn:backpropagation}), the first term on the right-hand side carries a negative sign, and under typical circumstances, the partial derivative with respect to \(\omega_{\alpha}\) maintains a consistent sign. However, for the specific class of activation functions discussed in this work, the third term on the right-hand side of Eq.~(\ref{eqn:neuralnetflow}) can estimate a negative sign within negative domains. This behavior is analaogous to the effect observed in the second line (i.e., the second arrow) of Eq.~(\ref{eqn:neuralnetflow}), where specific instances of \(y_{\alpha i}\) carry a negative sign, thereby inducing a perturbative influence during the update step.
Since the negative domain of the activation function is intended to introduce only a soft perturbative effect, it should take small negative values and rapidly decay toward zero as the input approaches negative infinity as well. This characteristic ensures that the influence of the negative region remains limited and non-disruptive, thereby contributing to both faster and more accurate learning performance in neural networks.

\section{Experiment}
The robustness of the proposed MoLU activation function lies in its ability to rapidly guide the optimization process toward the minimum of the loss function while maintaining numerical stability throughout training. As discussed earlier, MoLU is mathematically equivalent to the Swish with \(\beta=2\). However, interestingly, empirical results reveal that MoLU significantly outperforms Swish with \(\beta=2\), not only in terms of predictive performance, but also in terms of training speed. This unexpected advantage underscores the practical efficiency of MoLU and highlights the subtle importance of implementation dynamics and numerical behavior in deep learning optimization. To rigorously evaluate the performance of MoLU, we conducted a comprehensive set of experiments across several representative learning tasks. These include continuous dynamics modeling via Neural ODEs, as well as benchmark image classification tasks using MNIST and CIFAR-10 datasets. In the context of Neural ODEs, where differentiability of the activation function is a prerequisite, we compared MoLU against GeLU, SiLU, Mish, ELU, Tanh, and Swish with \(\beta=2\). For image classification tasks, MoLU was compared against ReLU, Leaky ReLU and Tanh, which are commonly used baseline activation functions. Our findings demonstrate that MoLU not only preserves training stability, but also consistently achieves faster convergence and superior accuracy across all evaluated domains.

\begin{figure*}[!hbt]
\centering
\includegraphics[width=1.0\textwidth]{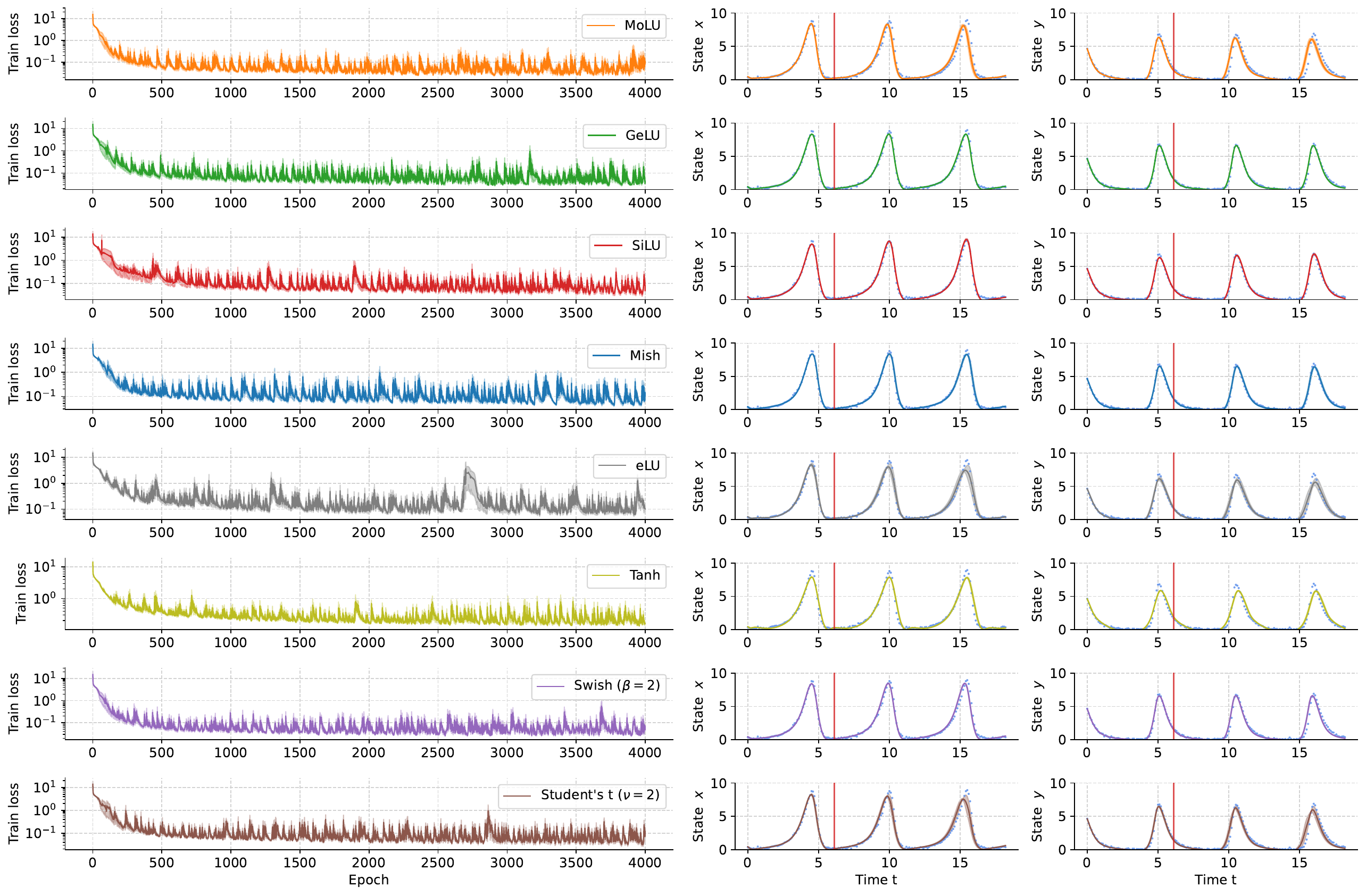}
\vskip -0.1in
\caption{Train losses. Train losses per epochs for each activations. Left region of vertical red line is training datasets and right region is extrapolation predictions.}
\label{fig:losses_states}
\end{figure*}

\FloatBarrier

\subsection{Neural ODEs}
\begin{figure}[!ht]
\centering
\includegraphics[width=0.45\textwidth]{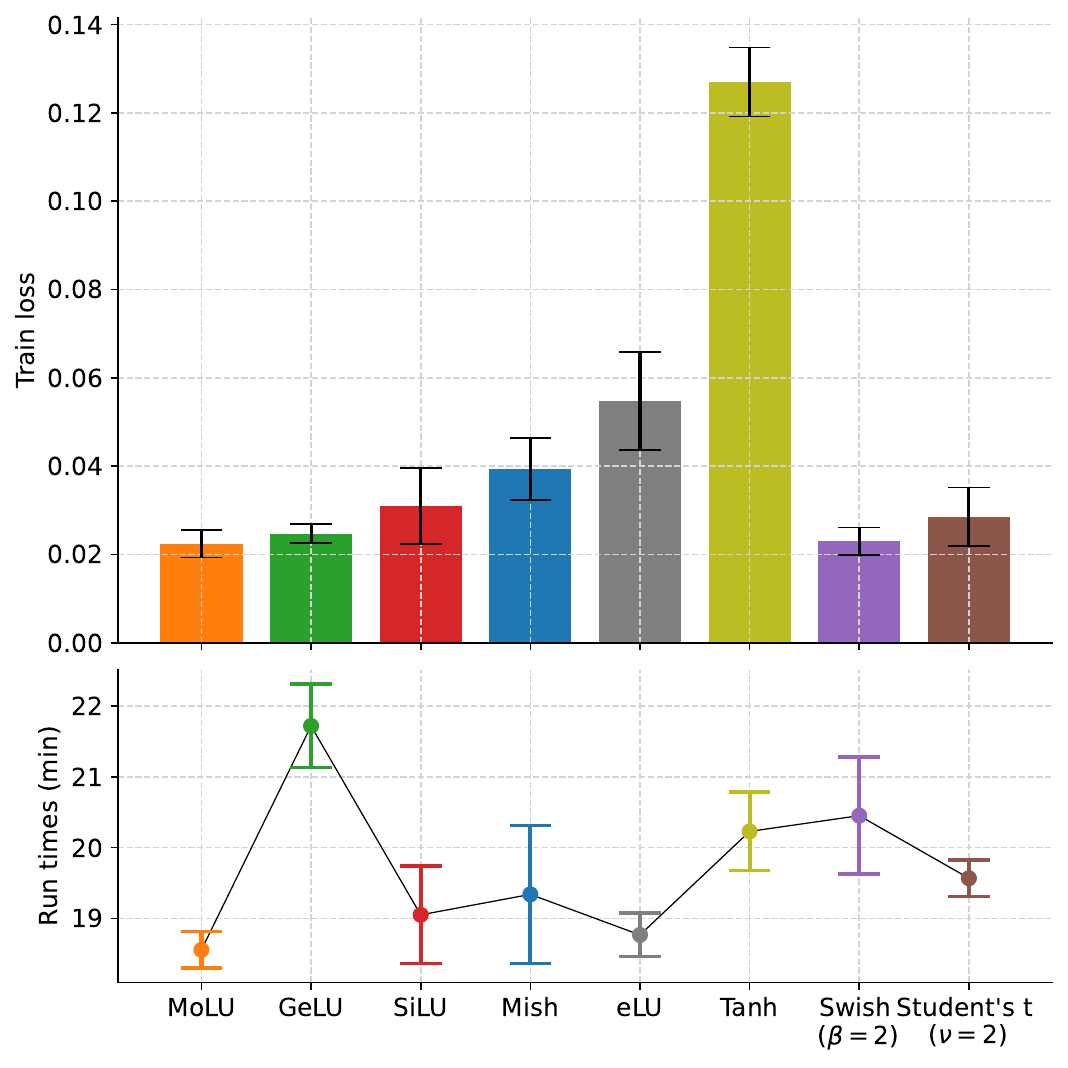}
\caption{Mean train losses with standard deviations. \textbf{(Top)}  Total training times extracted from WandB \cite{wandb}. \textbf{(Bottom)}}
\label{fig:bar}
\end{figure}

\vskip -0.1in
We begin our experiment with Lotka-Volterra model, the commonly used simple model of Neural ODEs. Coefficients and initial conditions in the Lotka-Volterra equations were all identical to the setting in \cite{homotopy}. Following \cite{stiff}, we generated training data by numerically over the time span $t \in [0, 6.1]$, then adding Gaussian noise with zero mean and standard deviation $5\%$ of the mean of each channel. We set 4,000 epochs for each experiment, the learning rate was 0.05, and random seeds were used for 10, 20, 30 for each activation function.\\
\cref{fig:losses_states} and \cref{fig:bar} show that our activation function is not only rapidly approaching the minimum of a loss function, but also showing a very stable performance.\\
It is worth noting that our implementation of MoLU was defined manually as a custom Python function, without any low-level optimization. In contrast, the other activation functions were invoked directly from torch.nn.functional, which are internally optimized for performance within the PyTorch framework. Despite this disadvantage, MoLU not only achieved the highest accuracy, but also exhibited the fastest training convergence among all tested activation functions.

\FloatBarrier

\begin{table}[hbt!]
\caption{Summary of mean training losses with standard deviations and training time for Neural ODEs using various activation functions.}
\begin{ruledtabular}
\begin{tabular}{l|cccccc}

\multicolumn{7}{c}{Mean train loss($10^{-2}$) / Standard dev.($10^{-3}$)} / Time (min) \\

    \hline
    
\multicolumn{1}{c}{} &  \multicolumn{1}{c}{\ MoLU} &    \multicolumn{1}{c}{\ GeLU} & \multicolumn{1}{c}{\ SiLU} & \multicolumn{1}{c}{\ Mish} & \multicolumn{1}{c}{\shortstack{\ Swish\\(\(\beta=2\))}} & \multicolumn{1}{c}{\shortstack{Student's t\\\((\nu=2)\)}} \\

    \hline

Train loss & \textbf{2.25} & 2.47 &\ 3.10 &\ 3.94 &\ \ 2.30 & 2.85 \\
Std. dev. & \ 3.13 & \textbf{2.10} &\ 8.67 &\ 7.01 &\ \ 3.08 & 6.67 \\
Train time & \ \textbf{18.6} & 21.7 &\ 19.1 &\ 19.3 &\ \ 20.5 & 19.6 \\

\end{tabular}
\end{ruledtabular}
\label{tab:NODE}
\end{table}

\subsection{MNIST}
\vskip -0.1in
We conducted experiments with MNIST, the most commonly used datasets in the field of image classification. We used MNIST datasets in torchvision \cite{torchvision} and used 2 layered Networks which is optimized using SGD on a batch size of 64 with a learning rate of 0.001 and a momentum of 0.5 with random seed of 10. We confirmed that our activation function shows a high-performance. Compared to other activation functions, our activation function clearly shows the characteristics of converging rapidly at the beginning of learning.

\begin{figure}[hbt!]
\centering
\includegraphics[width=0.45\textwidth]{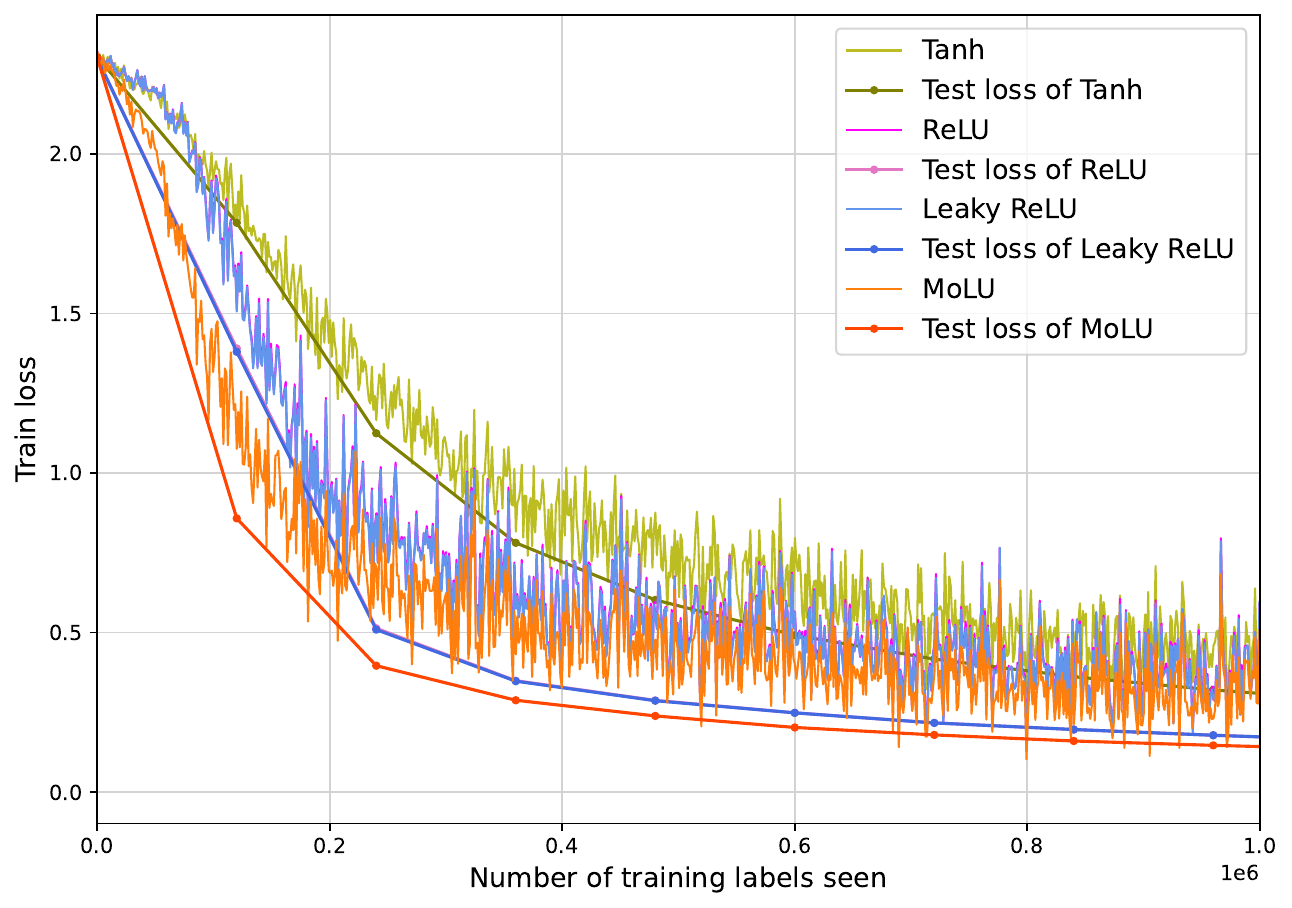}
\vskip -0.1in
\caption{MNIST Results. Average of loss.}
\label{fig:2_loss}
\end{figure}

% \begin{figure}[hbt!]
% \centering
% \includegraphics[width=0.4\textwidth]{figures/2,2,acc.pdf}
% \caption{\label{fig:1,1,acc}MNIST Results. Average of accuracy.}
% \end{figure}

\begin{table}[hbt!]
\caption{MNIST.Average of accuracy for some activation function.}
\begin{ruledtabular}
\begin{tabular}{l|cccccc}

\multicolumn{1}{c}{} & \multicolumn{4}{c}{MNIST (Accuracy(\%))} \\

    \hline
    
\multicolumn{1}{c}{} & \multicolumn{1}{c}{Tanh} & \multicolumn{1}{c}{ReLU} &  \multicolumn{1}{c}{Leaky ReLU} & \multicolumn{1}{c}{MoLU} \\

    \hline

1 ep. & 65.52 \% & 70.97 \% & 71.28 \% & \textbf{78.01} \% \\
2 ep. & 75.24 \% & 87.44 \% & 87.51 \% & \textbf{88.51} \% \\
3 ep. & 82.08 \% & 90.49 \% & 90.52 \% & \textbf{91.69} \% \\
4 ep. & 85.87 \% & 91.59 \% & 91.52 \% & \textbf{93.12} \% \\
5 ep. & 88.21 \% & 92.83 \% & 92.82 \% & \textbf{94.07} \% \\
10 ep. & 92.87 \% & 95.40 \% & 95.38 \% & \textbf{96.23} \% \\
20 ep. & 95.35 \% & 97.08 \% & 97.05 \% & \textbf{97.51} \% \\
30 ep. & 96.56 \% & 97.68 \% & 97.71 \% & \textbf{98.01} \% \\
\end{tabular}
\end{ruledtabular}
\label{tab:MNIST}
\end{table}

\FloatBarrier

\subsection{CIFAR10}
We conducted experiment on CIFAR10 which is more challenging model than MNIST in classification fields. ResNet18 \cite{resnet} which is optimized using SGD on a batch size of 32 with a learning rate of 0.001, momentum of 0.9 is used for the experiment with random seed of 10. Our activation function converges rapidly with respect to the Top-1 accuracy and the Top-5 accuracy in Table3, Table4.

% \vskip -0.3in

\begin{table}[hbt!]
\caption{Top-1 Accuracy. Top-1 accuracy of prediction on CIFAR10 for some activation function.}
\begin{ruledtabular}
\begin{tabular}{l|cccccc}

\multicolumn{1}{c}{} & \multicolumn{4}{c}{CIFAR10 (Top-1 Accuracy)} \\

    \hline
    
\multicolumn{1}{c}{} & \multicolumn{1}{c}{Tanh} & \multicolumn{1}{c}{ReLU} &  \multicolumn{1}{c}{Leaky ReLU} & \multicolumn{1}{c}{MoLU} \\

    \hline
1 ep. & 47.28 \% & 58.04 \% & \textbf{63.95} \% & 63.66 \% \\
2 ep. & 55.24 \% & 73.96 \% & \textbf{74.09} \% & 72.22 \% \\
3 ep. & 59.47 \% & \textbf{77.38} \% & 77.17 \% & 77.04 \% \\
4 ep. & 65.23 \% & 77.36 \% & 77.23 \% & \textbf{78.12} \% \\
5 ep. & 66.98 \% & 77.46 \% & \textbf{78.80} \% & 78.30 \% \\
10 ep. & 70.01 \% & 80.51 \% & \textbf{81.71} \% & 80.17 \% \\
20 ep. & 70.41 \% & 82.60 \% & \textbf{83.00} \% & 82.90 \% \\
30 ep. & 72.60 \% & 83.20 \% & \textbf{83.51} \% & 82.97 \% \\
\end{tabular}
\end{ruledtabular}
\label{tab:CIFAR10}
\end{table}

\begin{table}[hbt!]
\caption{Top-5 Accuracy. Top-5 accuracy of prediction on CIFAR10 for some activation function.}
\begin{ruledtabular}
\begin{tabular}{l|cccccc}

\multicolumn{1}{c}{} & \multicolumn{4}{c}{CIFAR10 (Top-5 Accuracy)} \\

    \hline
    
\multicolumn{1}{c}{} & \multicolumn{1}{c}{Tanh} & \multicolumn{1}{c}{ReLU} &  \multicolumn{1}{c}{Leaky ReLU} & \multicolumn{1}{c}{MoLU} \\

    \hline
1 ep. & 91.29 \% & 95.25 \% & \textbf{96.94} \% & 96.62 \% \\
2 ep. & 94.48 \% & 98.26 \% & \textbf{98.49} \% & 98.21 \% \\
3 ep. & 95.53 \% & 98.65 \% & 98.48 \% & \textbf{98.74} \% \\
4 ep. & 96.67 \% & \textbf{98.97} \% & 98.54 \% & 98.36 \% \\
5 ep. & 96.79 \% & 98.67 \% & \textbf{98.72} \% & 98.64 \% \\
10 ep. & 97.35 \% & 98.72 \% & \textbf{98.84} \% & 98.81 \% \\
20 ep. & 96.92 \% & 98.66 \% & 98.69 \% & \textbf{98.79} \% \\
30 ep. & 97.02 \% & 98.74 \% & 98.73 \% & \textbf{98.75} \% \\
\end{tabular}
\end{ruledtabular}
\label{tab:CIFAR10_5}
\end{table}

\section{Conclusion}
The MoLU activation function exhibited notably enhanced convergence behavior and training stability in Neural Ordinary Differential Equation (Neural ODE) and benchmark image classification tasks using
MNIST and CIFAR-10 datasets. While its mathematical form is equivalent to the Swish activation with \(\beta=2\), empirical results reveal that MoLU induces significantly different learning dynamics.

\begin{acknowledgments}
This work was supported by grants from the National Research Foundation of Korea (No. 2016R1A3B1908660) to W. Jhe.
% We wish to acknowledge the support of the author community in using
% REV\TeX{}, offering suggestions and encouragement, testing new versions,
% \dots.
\end{acknowledgments}

\FloatBarrier

\appendix

\section{Appendixes}

We further conducted experiments by modifying the scale of the negative domain (i.e., the region before convergence to zero), either compressing or stretching it. To preserve the linearity in the positive region, we retained the multiplicative modulation of the cumulative distribution function (CDF) by the input. However, we systematically varied the input to the CDF by either multiplying or dividing it by 2. Across activation functions, we ran experiments with random seeds 10, 20, and 30. The results consistently revealed that compressing the negative domain (i.e., input multiplied by 2) led to more effective learning than stretching it (i.e., input divided by 2).

\vskip 0.1in
\begin{enumerate}
    \item \textbf{Stretched the negative domain}
    \[
    A(x) = x \cdot \Phi(x/2).
    \]
    \item \textbf{Compressed the negative domain}
    \[
    A(x) = x \cdot \Phi(2x).
    \]
\end{enumerate}

\vskip -0.2in
\begin{figure}[!ht]
\centering
\includegraphics[width=0.4\textwidth]{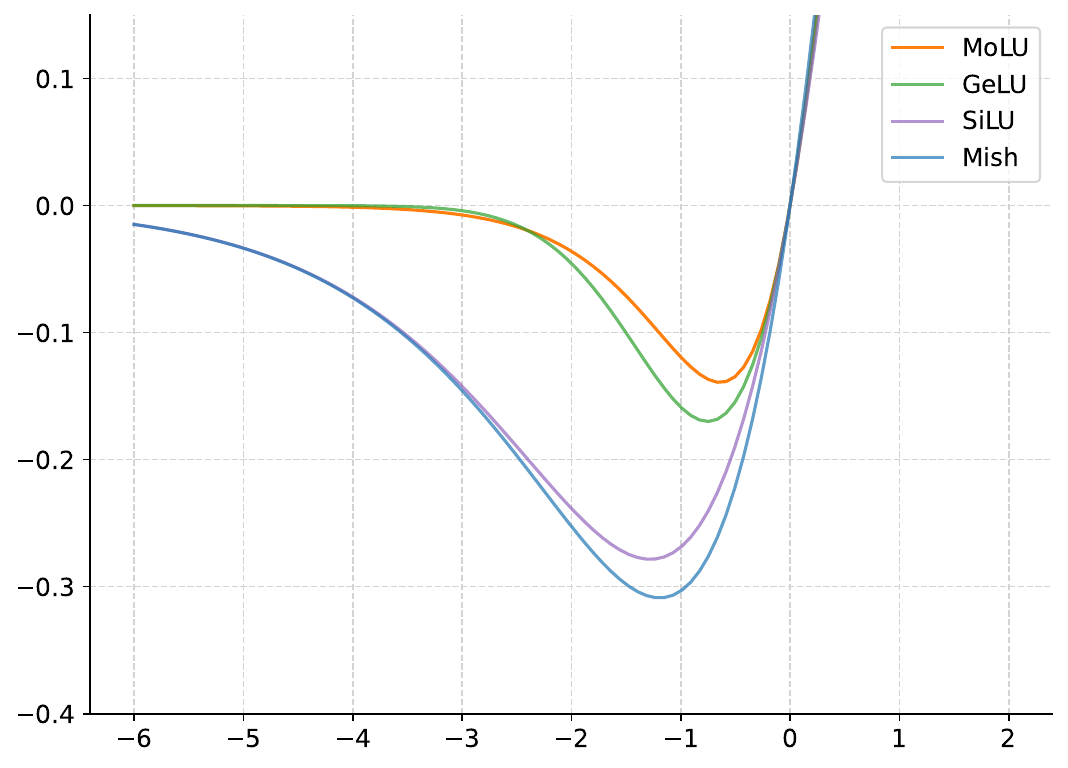}
\vskip -0.1in
\caption{Four activation functions for the experiment.}
\label{fig:act4}
\end{figure}

\vskip -0.1in
\begin{figure}[!ht]
\centering
\includegraphics[width=0.4\textwidth]{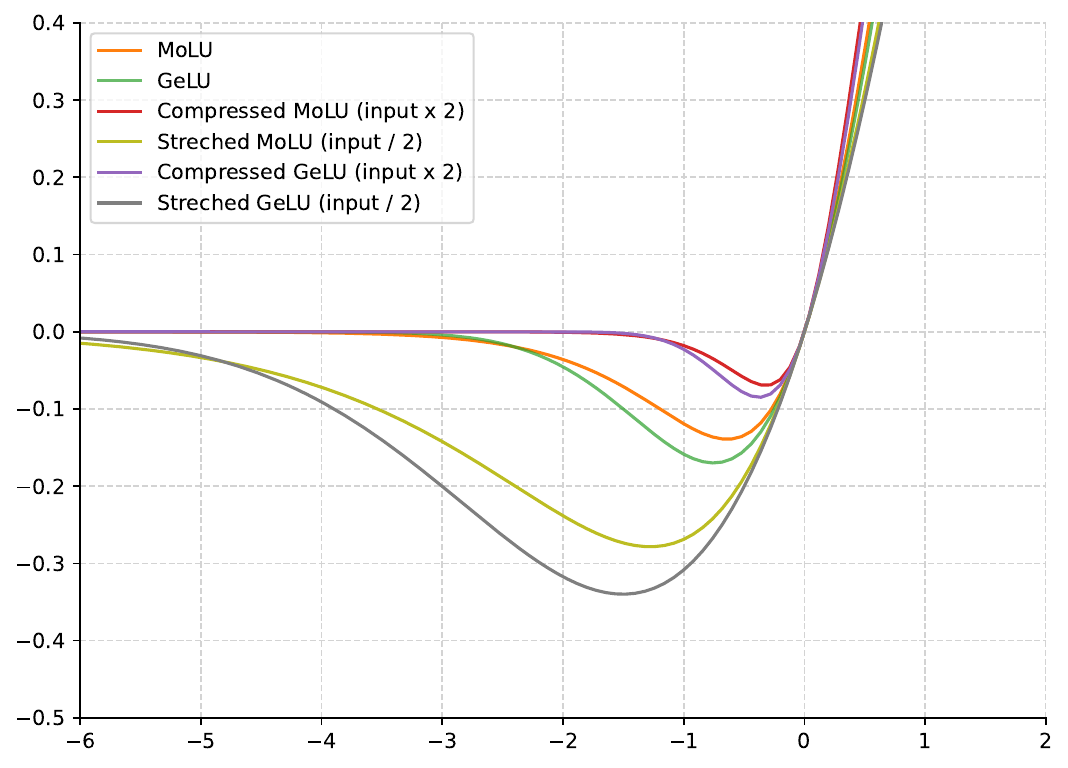}
\vskip -0.1in
\caption{Two activation functions with a compressed and stretched negative domain.}
\label{fig:act_}
\end{figure}

\vskip -0.1in
\begin{figure}[!ht]
\centering
\includegraphics[width=0.4\textwidth]{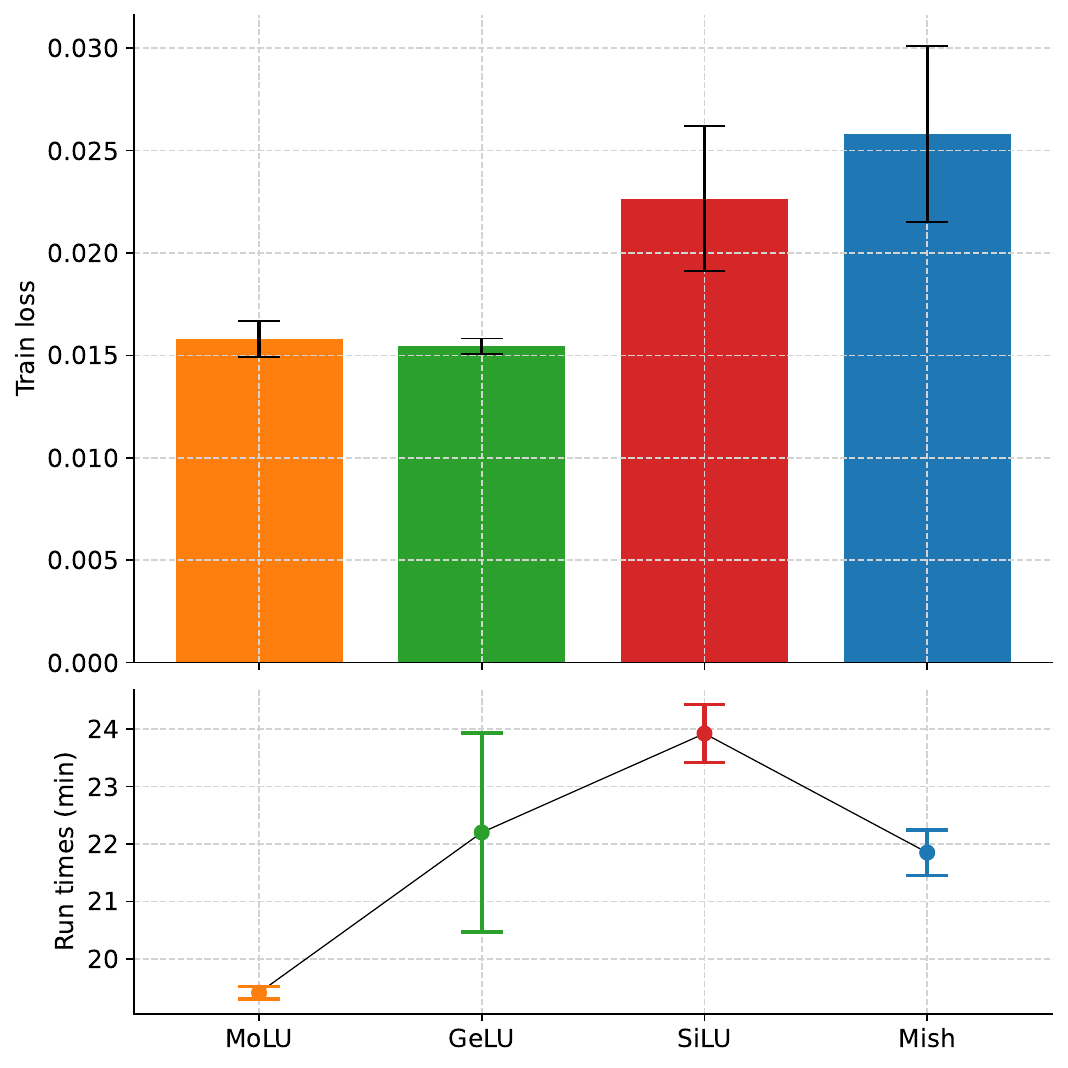}
\vskip -0.1in
\caption{Training losses for activation functions with \textbf{a compressed negative domain}.}
\label{fig:bar_c}
\end{figure}

\vskip -0.1in
\begin{figure}[!ht]
\centering
\includegraphics[width=0.4\textwidth]{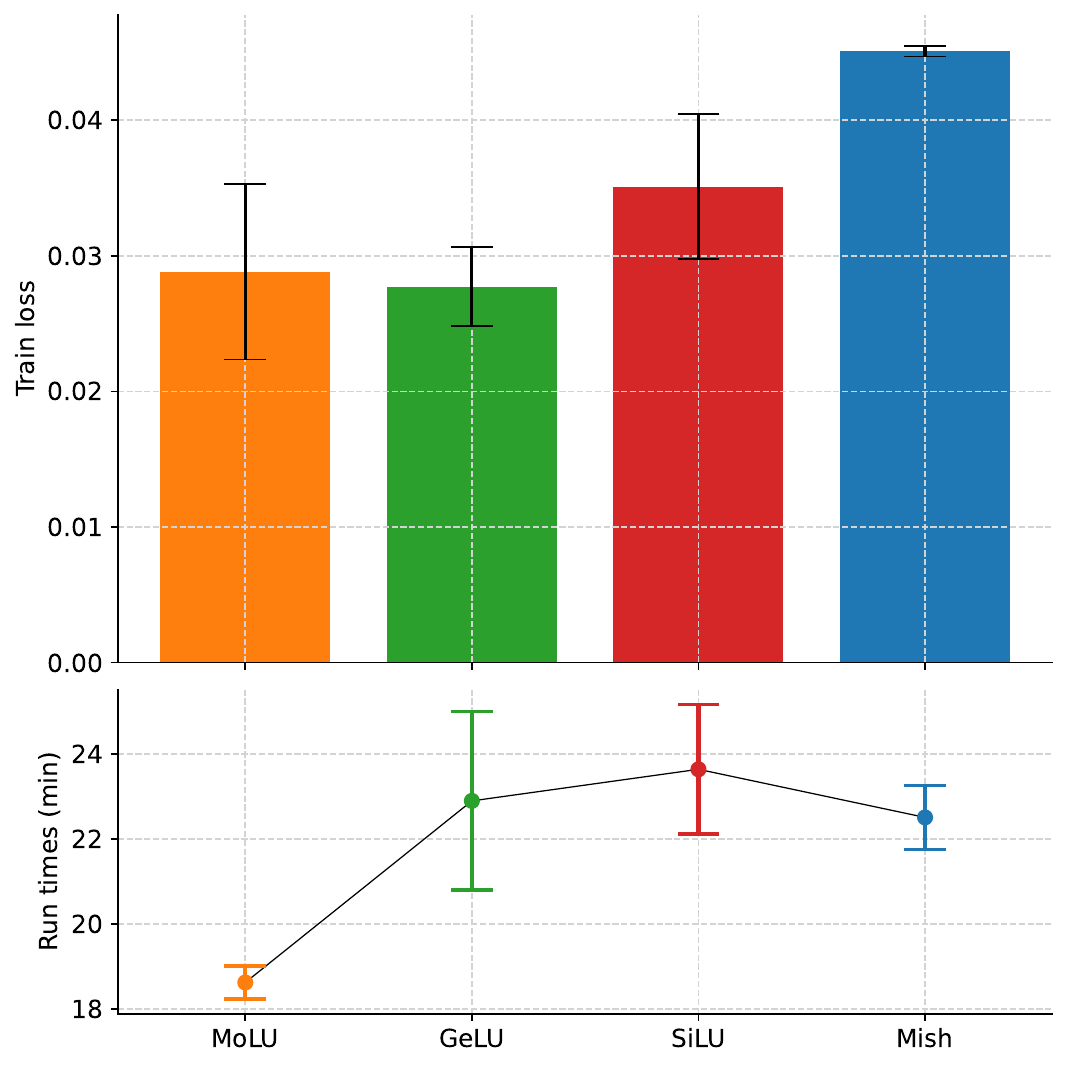}
\vskip -0.1in
\caption{Training losses for activation functions with \textbf{a stretched negative domain}.}
\label{fig:bar_s}
\end{figure}

\FloatBarrier

\vskip -0.3in

% \begin{table}[ht]
% \centering
% \caption{Mean training losses, standard deviations, and training times for activation functions with a compressed (top three rows) and stretched (bottom three rows) negative domain.}
% \begin{tabularx}{\textwidth}{>{\raggedright\arraybackslash}X *{4}{>{\centering\arraybackslash}X}}
% \toprule
% & MoLU & GeLU & SiLU & Mish \\
% \midrule
% Train loss ($10^{-2}$) & 1.58 & \textbf{1.54} & 2.26 & 2.58 \\
% Std. dev. ($10^{-3}$)  & 0.88 & \textbf{0.39} & 3.54 & 4.29 \\
% Train time (min)       & \textbf{19.4} & 22.2 & 23.9 & 21.85 \\
% \midrule
% Train loss ($10^{-2}$) & 2.88 & \textbf{2.77} & 3.51 & 4.51 \\
% Std. dev. ($10^{-3}$)  & 6.45 & 2.91 & 5.33 & \textbf{0.39} \\
% Train time (min)       & \textbf{18.6} & 22.9 & 23.6 & 22.5 \\
% \bottomrule
% \end{tabularx}
% \label{tab:app}
% \end{table}

% \begin{table}[ht]
% \caption{Mean training losses, standard deviations, and training times for activation functions with a compressed (top three rows) and stretched (bottom three rows) negative domain.}
% \begin{tabularx}{\columnwidth}{Y Y Y Y Y}
% \toprule
%  & MoLU & GeLU & SiLU & Mish \\
% \midrule
% Train loss ($10^{-2}$) & 1.58 & \textbf{1.54} & 2.26 & 2.58 \\
% Std. dev. ($10^{-3}$)  & 0.88 & \textbf{0.39} & 3.54 & 4.29 \\
% Train time (min)       & \textbf{19.4} & 22.2 & 23.9 & 21.9 \\
% \midrule
% Train loss ($10^{-2}$) & 2.88 & \textbf{2.77} & 3.51 & 4.51 \\
% Std. dev. ($10^{-3}$)  & 6.45 & 2.91 & 5.33 & \textbf{0.39} \\
% Train time (min)       & \textbf{18.6} & 22.9 & 23.6 & 22.5 \\
% \bottomrule
% \end{tabularx}
% \label{tab:activation_results}
% \end{table}

\begin{table}[hbt!]
\caption{Summary for the activation functions with a compressed (\textbf{upper three rows}) and stretched  (\textbf{lower three rows}) negative domain.}
\begin{ruledtabular}
\begin{tabular}{p{1.8cm} C{0.8cm} C{0.8cm} C{0.8cm} C{0.8cm}}

\multicolumn{5}{c}{Train loss ($10^{-2}$) / Std.($10^{-3}$)}Time (min)\\

    \hline
    
\multicolumn{1}{c}{} & \multicolumn{1}{c}{MoLU} & \multicolumn{1}{c}{GeLU} & \multicolumn{1}{c}{SiLU} & \multicolumn{1}{c}{Mish}\\

    \hline

Train loss & 1.58 & \textbf{1.54} & 2.26 & 2.58 \\
Std. dev. & 0.88 & \textbf{0.39} & 3.54 & 4.29 \\
Train time & \textbf{19.4} & 22.2 & 23.9 & 21.9 \\ 
    \hline
Train loss & 2.88 & \textbf{2.77} & 3.51 & 4.51 \\
Std. dev. & 6.45 & 2.91 &\ 5.33 & \textbf{0.39} \\
Train time & \textbf{18.6} & 22.9 & 23.6 & 22.5 \\ 

\end{tabular}
\end{ruledtabular}
\label{tab:app}
\end{table}

\bibliography{references}% Produces the bibliography via BibTeX.

\end{document}